\documentclass[sigconf]{aamas} 

\usepackage{balance} 

\usepackage{times}
\usepackage{helvet}
\usepackage{courier}
\usepackage{amsmath}
\usepackage{graphicx}
\usepackage{amsbsy}
\usepackage{bbm}
\usepackage{xcolor}
\usepackage{subcaption}

\RequirePackage{algorithm}
\RequirePackage{algcompatible}
\RequirePackage{algpseudocode}

\usepackage{amsfonts}
\usepackage{booktabs}
\usepackage{siunitx}

\usepackage{tabularx,booktabs,caption}
\newcolumntype{Z}{>{\raggedright}X}

\definecolor{Orange}{rgb}{1,0.5,0}
\definecolor{Red}{rgb}{1,0,0}
\definecolor{Blue}{rgb}{0,0,1}

\setcopyright{ifaamas}
\acmConference[]{}{}
{}{}
\copyrightyear{2023}
\acmYear{2023}
\acmDOI{}
\acmPrice{}
\acmISBN{}

\title[AAMAS-2023 Formatting Instructions]{ XDQN: Inherently Interpretable DQN through Mimicking}

\author{Andreas Kontogiannis}
\affiliation{
  \institution{National Technical University of Athens}
  \country{Greece}
    }
\email{andr.kontog@gmail.com}

\author{George Vouros}
\affiliation{
  \institution{University of Piraeus}
  \country{Greece}
  }
\email{georgev@unipi.gr}

\begin{abstract}
Although deep reinforcement learning (DRL) methods have been successfully applied in challenging tasks, their application in real-world operational settings is challenged by methods' limited ability to provide explanations. Among the paradigms for explainability in DRL is the interpretable box design paradigm, where interpretable models substitute inner constituent models of the DRL method, thus making the DRL method “inherently" interpretable. In this paper we explore this paradigm and we propose XDQN, an explainable variation of DQN, which uses an interpretable policy model trained through mimicking. XDQN is challenged in a complex, real-world operational multi-agent problem, where agents are independent learners solving congestion problems. Specifically, XDQN is evaluated in three MARL scenarios, pertaining to the demand-capacity balancing problem of air traffic management. XDQN achieves performance similar to that of DQN, while its abilities to provide global models' interpretations and interpretations of local decisions are demonstrated.
\end{abstract}

\keywords{Deep Reinforcement Learning, Mimic Learning, Explainability}

\newcommand{\BibTeX}{\rm B\kern-.05em{\sc i\kern-.025em b}\kern-.08em\TeX}

\begin{document}

\pagestyle{fancy}
\fancyhead{}


\maketitle 

\section{Introduction}

Deep Reinforcement Learning (DRL) has mastered decision making policies in various difficult control tasks \cite{7989385} \cite{Levy2019LearningMH} \cite{Kravaris2022ExplainingDR}, games \cite{mnih2015humanlevel} \cite{10.5555/3016100.3016191} and other real-time applications \cite{Kontogiannis2021TreebasedFW} \cite{Zhao_Gu_Zhang_Yang_Liu_Tang_Liu_2021}. Despite the remarkable performance of DRL models, the knowledge of mastering these tasks remains implicit in deep neural networks. Thus, its application in real-world operational settings is challenged by methods’ limited ability to provide explanations at global (policy) and local (individual decisions) levels. This lack of interpretability makes it difficult to trust DRL for solving safety-critical real-world tasks. However, besides the inability of DRL models to provide interpretations on the selection of actions in specific circumstances, they are also unable to provide information about the evolution of models during the training process. These challenges are naturally further extended to multi-agent settings, in which different agents empowered by multi-agent reinforcement learning (MARL) methods aim at learning a joint optimal policy towards solving a target task.

To address some of the aforementioned challenges, one may follow different paradigms for the provision of explanations: The interpretable box design paradigm is one of them where interpretable models substitute inner components of DRL \cite{10.1145/3527448}. Additionally, mimic learning has been proposed, so as to infer interpretable models that mimic the behavior of well-trained deep neural networks \cite{NIPS2014_ea8fcd92, 10.1145/775047.775113}. In the DRL case, mimic learning aims to replace the closed-box DRL controller with an interpretable one, able to mimic the decisions made by the former \cite{10.1145/3527448, Liu2018TowardID, 10.5555/3327144.3327175}. A mimic learner tries to optimize \textit{fidelity} \cite{10.1145/3527448}, which is determined by comparing the mimic controller's actions with the actions selected by the DRL model. To extract knowledge from deep neural networks, recent work \cite{Liu2018TowardID, 10.5555/3327144.3327175} has applied mimic learning with tree representations, using decision trees: Criteria used for splitting tree nodes provide a tractable way to explain the predictions made by the controller. 

Typically, mimic learning approaches require already well-trained complex policy networks (which we refer to as \textit{mature} networks), whose behavior are mimicking to support interpretability. In real-world scenarios, this could be quite impractical, since the training overhead required to train the mimic models can often be a very time-consuming and costly process, especially for large state-action spaces and for multi-agent settings. Another limitation of such approaches is that they solely aim at providing explainability on the predictions of only the mature DRL model, ignoring completely the training process of this model. In other words, in these approaches, the mimic learner can only provide explanations about the policy of the inferred DRL controller, but not about the patterns and behaviors learned throughout the training process. 

To deal with these challenges, in this paper we propose \textit{e\textbf{X}plainable \textbf{D}eep \textbf{Q}-\textbf{N}etwork} (\textit{\textbf{XDQN}}), which is an explainable variation of the well-known DQN \cite{mnih2015humanlevel} method. In XDQN, our goal is to provide {inherent explainability of DQN via mimic learning} in an online manner, by replacing the complex deep Q-network with an interpretable mimic learner in testing. {In so doing, XDQN does not require the existence of a well-trained model to train an interpretable one}. In particular, we train a mimic learner in parallel with the deep neural network (Q-network) of DQN in an online setting, where: at a training step the DRL model uses the mimic learner to compute the target values of the Q-network needed for its training, while the mimic learner learns to behave as the DRL model, but in an explainable way. {Since the mimic learner is trained and updated while the DQN policy model is trained, we can keep multiple ``snapshots'' of the model evolution through time, offering interpretability on these intermediate models, and insights about the patterns and behaviors that DQN learns during training.}

To evaluate our method's utility in real-world operational settings, XDQN is challenged in a complex, real-world multi-agent problem, where
agents solve airspace congestion problems. Agents in this setting are trained via parameter sharing following the centralized training, decentralized execution paradigm. We summarize the main contributions of this paper below:

\begin{itemize}
    \item To our knowledge, this work is the first that {provides DQN with inherent interpretability through mimic learning without requiring} the existence of a  well-trained DRL model.
    \item We propose XDQN, an explainable variation of DQN, in which an interpretable mimic learner is trained in parallel with the Q-network of DQN and plays the role of the target Q-network of DQN.
    \item Experimentally, we show that XDQN can perform similarly to DQN, demonstrating good play performance and fidelity to DQN decisions in complex, real-world operational multi-agent problems.
    \item We demonstrate the ability of XDQN to provide global (policy) and local (in specific circumstances) explanations regarding agents' decisions, {also while models are being trained}.
\end{itemize}


\section{Background}

\subsection{Markov Decision Process}

We consider a sequential decision making setup, in which an agent interacts with an environment $E$ over discrete time steps. At a given timestep, the agent perceives features regarding a state $s_t \in S$, where $S$ is the state space. The agent then chooses an action $a_t$ from a discrete set $A$ and observes a reward $r_t$ generated by the environment.

The agent's behavior is determined by a policy $\pi$, which maps states to a probability distribution over the actions, that is $\pi \colon S \rightarrow P(A)$. Apart from an agent's policy, the environment $E$ may also be stochastic. We model it as a Markov
Decision Process (MDP) with a state space $S$, action space $A$, an initial state distribution $p(s_1)$, transition dynamics $p(s_{t+1}|s_t)$ and a reward function $r(s_t, a_t, s_{t+1})$. For brevity, we write $r_t = r(s_t, a_t, s_{t+1})$. 

The agent aims to maximize the expected discounted cumulative reward, which is formulated as $G_t = \sum_{\tau=t}^{\infty}\gamma^{\tau-t}r_{\tau}$.
Here, $\gamma \in (0,1)$ is a discount factor which trades-off the importance of immediate and future rewards. Considering that an agent acts under a stochastic policy $\pi$, the Q-function (state-action value) of a pair $(s,a)$ is defined as follows

\begin{equation}
Q^{\pi}(s,a) = \mathbb{E}\left[G_t \mid s_t=s, a_t=a, \pi \right]
\end{equation}

\noindent
which can also be computed recursively with bootstrapping:

\begin{equation}
Q^{\pi}(s,a) = \mathbb{E}\left[r_t + \gamma \mathbb{E}_{a \sim \pi(s_{t+1})}[Q^{\pi}(s_{t+1},a)] \mid s_t=s, a_t=a, \pi \right]
\end{equation}

\noindent
The Q-function measures the value
of choosing a particular action when the agent is in this state. We define the optimal policy $\pi^*$ under which the agent receives the optimal $Q^*(s,a) = max_{\pi}Q^{\pi}(s,a)$. For a given state $s$, under the optimal policy $\pi^*$, the agent selects action $a = argmax_{a' \in A} Q^*(s,a')$. Therefore, it follows that the optimal Q-function satisfies the Bellman equation:

\begin{equation}
Q^*(s,a) = \mathbb{E}\left[r_t + \gamma max_{a} Q^*(s_{t+1},a) \mid s_t=s, a_t=a, \pi \right] \text{.}
\end{equation} 

\subsection{Deep Q-Networks}
To deal with a high dimensional state space, the state-action value function can be approximated by an \textit{online} deep Q-network (DQN \cite{mnih2015humanlevel}); i.e. a deep neural network $Q(s,a;\theta)$ with weight parameters $\theta$. To estimate the parameters $\theta$, at iteration $i$  the expected mean squared loss between the estimated Q-value of a state-action pair and its temporal difference target, produced by a fixed and separate \textit{target} Q-network $Q(s,a;\theta^{-})$ with weight parameters $\theta^{-}$, is minimized. Formally:

\begin{equation}
L_i(\theta_i) = \mathbb{E}\left[Y_i^{DQN} - Q(s,a;\theta)\right] \text{,} \\
\end{equation} 
with
\begin{equation}
Y_i^{DQN} = r_{t} + \gamma \max_{a \in A} Q(s_{t+1},a;\theta^-)
\end{equation} 

\medskip

\noindent 
In order to train DQN and estimate $\theta$, we could use the standard Q-learning update algorithm. Nevertheless, the Q-learning estimator performs very poorly in practice. To stabilize the training procedure of DQN, Mnih et. al \cite{mnih2015humanlevel} freezed the parameters, $\theta^-$, of the target Q-network for a fixed number of training iterations while updating the online Q-network with gradient descent steps with respect to $\theta$. 


In addition to the target network, during the learning process, DQN uses an experience replay buffer \cite{mnih2015humanlevel}, which is an accumulative dataset, $D_t$, of  state transitions - in the form of ($s$, $a$, $r$, $s'$) - from past episodes. In a training step, instead of only using the current state transition, the Q-Network is trained by sampling mini-batches of past transitions from $D$ uniformly, at random. Therefore, the loss can be written as follows:

\begin{equation}
L_i(\theta_i) = \mathbb{E}_{(s,a,r,s') \sim \textit{U}(D)}\left[(Y_i^{DQN} - Q(s,a;\theta))^2\right] \text{.}
\end{equation} 

\noindent
The main advantage of using an experience replay buffer is that uniform sampling  reduces the correlation among the experience samples used for training the Q-network. The replay buffer also improves data efficiency through reusing the experience samples in multiple training steps.




Instead of sampling mini-batches of past transitions uniformly from the experience replay buffer, a further improvement over DQN results from using a prioritized experience replay buffer \cite{https://doi.org/10.48550/arxiv.1511.05952}. It aims at increasing the probability of sampling those past transitions from the experience replay that are expected to be more useful in terms of absolute temporal difference error. 

\subsection{Mimic Learning for Deep Reinforcement Learning}
Recent work on mimic learning \cite{Che2016InterpretableDM, Liu2018TowardID} has shown that rule-based models, like decision trees, or shallow feed-forward neural networks can mimic a not linear function inferred by a deep neural network with millions of parameters. 
We present two known settings for mimicking the Q-function of a DRL model.

\subsubsection{Experience Training} 

In the experience training setting \cite{Che2016InterpretableDM, Liu2018TowardID}, all the state-action pairs $\langle s,a \rangle$ of a DRL training process are collected in a time horizon $T$. Then, to obtain the corresponding Q-values, these pairs are provided as input into a DRL model. The final set $\{(\langle s_1,a_1\rangle, {Q}_1), ... (\langle s_T,a_T\rangle, {Q}_T)\}$ of tuples is used as the experience training dataset.

\subsubsection{Active Play}

The main problem with the experience training is that suboptimal state-action pairs are collected through training, making it more difficult for a learner to mimic the behavior of the DRL model. To address this challenge, active play  \cite{Liu2018TowardID} uses a mature DRL model to generate state-action pairs to construct the training dataset of  an active mimic learner. The training data is collected in an online manner through queries, in which the active learner selects the actions, given the states, and the mature DRL model provides the estimated Q-values. These Q-values are then used to update the active learner's parameters on minibatches of the collected dataset. 

\section{Explainable Deep Q-Network (XDQN)}


\begin{algorithm*}[h!]
\caption{eXplainable Deep Q-Network (XDQN)}\label{alg:1}
\begin{algorithmic}[1]
    \State Initialize replay buffer $B$ with capacity N
    \State Initialize $\theta$ and $\phi_X$
    \State Initialize timestep count $c = 0$
    \For {episode 1, M} 
        \State Augment $c = c + 1$
        \State Initialize state $s_1$
        \State With probability $\epsilon$ select a random action $a_t$, otherwise $a_t = argmax_{a}Q(s_t,a;\theta)$
        \State Execute action $a_t$ and observe next state $s_{t+1}$ and reward $r_t$
        \State Store transition $(s_t, a_t, s_{t+1}, r_t)$ in $B$
        \State Sample a minibatch of transitions $(s_i, a_i, s_{i+1}, r_i)$ from $B$
        \If{$s_{i+1}$ not terminal} 
            \State Set $Y_i^{XDQN} = r_{i} + \gamma \max_{a \in A} Q\left(s_{i+1},a; \phi_X\right)$
        \Else
            \State Set $Y_i^{XDQN} = r_{i}$
        \EndIf
        \State Perform a gradient descent step on $\left(Y_i^{XDQN} - Q(s_i,a_i;\theta)\right)^2$ w.r.t. $\theta$
        
        \If{$c$ mod $T_u = 0$}
            \State Initialize $\phi_X$ 
            \State Sample a minibatch of transitions $(s_i, a_i, s_{i+1}, r_i)$ from $B$ that were stored at most $c-K$ timesteps before
            \State Perform mimic learning update on $\left( Q(s,a;\phi_X) - Q(s,a;\theta) \right)^2$ w.r.t $\phi_X$
        \EndIf
    \EndFor
\end{algorithmic}
\end{algorithm*}

In this work, we are interested in providing interpretability in deep Q-learning through mimicking the behavior of DQN. To this aim, we propose \textit{eXplainable Deep Q-Network} (\textit{XDQN}) \footnote{The implementation code will be made available in the final version of the manuscript.}, which is an explainable variation of DQN \cite{mnih2015humanlevel}. XDQN aims at  inferring the parameters of the online Q-network and the parameters of a mimic learner concurrently, in an online manner, with the latter substituting the target Q-network of DQN.

Formally, let $\theta$ be the parameters of the online Q-network and $\phi_X$ be the parameters of the mimic learner. In XDQN, the mimic learner is used to  estimate the state-action value function and {select the best action for the next state} in the XDQN target: 

\begin{equation}
Y_i^{XDQN} = r_{t} + \gamma \max_{a \in A} Q\left(s_{t+1},a; \phi_X\right)
\end{equation} 

\noindent

\noindent
Similar to DQN, $\phi_X$ are updated  every $T_u$ number of timesteps. The full training procedure of XDQN is presented in Algorithm \ref{alg:1}.

In contrast to DQN in which we simply copy the parameters $\theta$ of the online Q-network to update the parameters of the target Q-network, here we perform mimic learning on $Q(s,a,\theta)$ (steps 17-20). To update $\phi_X$ we train the mimic learner on minibatches of the experience replay buffer $B$ by minimizing the Mean Squared Error (MSE) loss function using $Q(s,a,\theta)$ to estimate the soft labels (Q-values) of the state-action pairs in the minibatches. Formally the optimization problem for each update of $\phi_X$ can be written as:

\begin{equation}
\min_{\phi_X} \mathbb{E}_{(s,a) \sim B} \left[ \left( Q(s,a;\phi_X) - Q(s,a;\theta) \right)^2  \right]
\end{equation} 

In our experiments, we utilize a prioritized experience replay \cite{https://doi.org/10.48550/arxiv.1511.05952} as the replay buffer $B$, as described in Section 2. Similarly to active play, when updating $\phi_X$, to ensure that the state-action pairs of the minibatches provide up-to-date target values with respect to $\theta$, we use records from the replay buffer that were stored during the $K$ latest training steps. 

It is worth noting that at each update of $\phi_X$ the hyperparameter $K$ for past transitions plays a similar role as the discounted factor $\gamma$ plays for future rewards, but from the mimic learner's perspective. Building upon the experience training and active play paradigms, XDQN can leverage the benefits of both of them. In particular, the hyperparameter $K$ manages the trade-off between experience training and active play in XDQN. If $K$ is large, the mimic model learns from state-action pairs that may have been collected through more suboptimal instances of $\theta$; deploying however data-augmented versions of Q-value. On the other hand, if $K$ is small, it learns from the most recent instances of $\theta$; making use of up-to-date Q-values. Nevertheless, opting for very small values of $K$ could lead to less stable mimic training, due to the smaller number of minibatches that can be produced for updating $\phi_X$, {while using large $K$ can result in a very slow training process}. 

From all the above, we note that $\theta$ (Q-network) and $\phi_X$ (mimic learner) are highly dependent. To update $\theta$, Q-network uses the mimic learner model with $\phi_X$  to compute the target soft labels (target Q-values), while to update $\phi_X$ the mimic learner uses the original Q-network with parameters $\theta$ to compute the respective target soft labels (online Q-values). Since XDQN produces different instances of $\phi_X$ throughout training, it can eventually output multiple interpretable mimic learner models (up to the number of $\phi_X$ updates), with each one of them corresponding to a different training timestep. Assuming that all these mimic learner instances are interpretable models, XDQN can also provide explainability on how a DRL model learns to solve the target task.

Finally, after Q-network ($\theta$) and  mimic learner ($\phi_X$) have been trained, without requiring to learn $\theta$ before $\phi_X$, we can discard the online Q-network and use the mimic learner model as the controller. Therefore, in testing, given a state, the interpretable mimic learner selects the action that profits the highest Q-value, being also able to provide explainability. 

\section{Experimental Setup}

In this section, we demonstrate the effectiveness of XDQN through experiments on real-world data. In all experiments we utilize a Gradient Boosting Regressor \cite{10.5555/3008751.3008849} as the mimic learner, so as to exploit its boosting ability to learn effectively by exploiting instances generated by the deep Q-network. Although most decision tree algorithms, being rule-based models, are naturally interpretable models \cite{Liu2018TowardID, 10.5555/3327144.3327175}, this is not the case for a Gradient Boosting Regressor, since the boosting structure makes it very difficult to provide explainability. However, following the work in \cite{DELGADOPANADERO2022199}, we are able to enrich the Gradient Boosting Regressor mimic learner with the ability to provide explainability as follows: Given a state-action pair as an input of the mimic learner, we can measure the contribution of each state feature to the predicted Q-value. Therefore, our mimic learner is expected not only to mimic effectively the behavior of the DRL controller, but also, to give local and global explanations on its decisions.

Overall, we are interested in comparing the performance of XDQN with that of DQN in real-world environments where the latter has been state-of-the-art, and also designing appropriate experimental setups, aiming at studying XDQN interpretability. 
In so doing, we evaluate XDQN on  real-world operational multi-agent experimental scenarios, pertaining to the demand-capacity balancing (DCB) problem of air traffic management (ATM), which we describe next.

\subsection{Real-world demand-capacity problem setting}

The current ATM system is based on time-based operations resulting in DCB \cite{10.1007/978-3-319-64798-2_15} problems. To solve the DCB issues at the pre-tactical stage of operations, the ATM system opts for  methods that generate delays and costs for the entire system. In ATM, the airspace consists of a set of 3D {sectors} where each one these is characterized by a specific capacity. This is the number of flights that cross the sector during a specific period (e.g. of 20 minutes). The main challenge of dealing with the DCB problem in ATM is to reduce the number of cases where the demand of airspace use exceeds its capacity. These cases are called \textit{hotspots}. 

Recent work has transformed the DCB challenge to a multi-agent RL problem by formulating the setting as a multi-agent MDP \cite{10.1007/978-3-319-64798-2_15}. We follow the work and the experimental setup of \cite{10.1007/978-3-319-64798-2_15, 10.1145/3200947.3201010, article, Kravaris2019ResolvingCI, Kravaris2022ExplainingDR} and encourage the reader to see the problem formulation \cite{10.1007/978-3-319-64798-2_15} in details. In this setting, we consider a society of agents, where each agent is a flight (related to a specific aircraft) that needs to coordinate its decisions, so as to resolve hotspots that occur, jointly with other society agents. Agents' local states comprise 81 state variables related to: (a) the delay (in the range of $0,...,\text{maxDelay}$) set by the referring agent, (b) the number of hotspots in which the agent is involved in, (c) the sectors that it crosses, (d) the minutes that the agent is within each sector it crosses, (e) the periods in which the agent joins in hotspots in sectors, and (f) the minute of the day that the agent takes off. The tuple containing all agents' local states is the joint global state. Q-learning \cite{Tan1993MultiAgentRL} agents has been shown to achieve remarkable performance on this task \cite{Kravaris2022ExplainingDR}. In our experiments, all agents share parameters and replay buffer and act independently.

{A DCB scenario comprises multiple flights crossing various airspace sectors in a time horizon of 24h. This time horizon is segregated into simulation time steps}. At each simulation time step {(equal to 10 minutes of real time)}, given only the local state, each agent selects an action which is related to its preference to add ground delay regulating its flight, {in order to resolve hotspots in which it participates}. The set of local actions for each agent contains $|\text{maxDelay}+1|$  actions, at each simulation time step. We use $\text{maxDelay} = 10$. The joint (global) action is a tuple of local actions selected by the agents. Similarly, we consider local rewards and joint (global) rewards. The local reward is related to the cost per minute within a hotspot, the total duration of the flight (agent) in hotspots as well as to the delay that a flight has accumulated up to the simulation timestep \cite{Kravaris2022ExplainingDR}. 


\begin{table*}[t]
\setlength\tabcolsep{4.8pt} 
\begin{tabular*}{\textwidth}{ccccccc}
\toprule 
Scenario & \multicolumn{3}{c}{DQN} & \multicolumn{3}{c}{XDQN}\\
\cmidrule(lr){2-4} \cmidrule(l){5-7}
& Final Hotspots & Average Delay & Delayed Flights & Final Hotspots & Average Delay & Delayed Flights \\
\midrule 
20190705    & 38.4 & 13.04 & 1556.5 & 39.0 & 13.19 & 1618.54 \\ 
20190708    & 4.6 & 11.4 & 1387.2 & 6.0 & 11.73 & 1331.58   \\
20190714    & 4.8 & 10.72 & 1645.2 & 7.0 & 13.46 & 1849.49 \\
\bottomrule 
\end{tabular*}
\caption{Comparison of testing performance of DQN and XDQN on the three experimental ATM scenarios}\label{tab:table3}
\end{table*}

\subsection{Evaluation Metrics and Methods}
For the evaluation of the proposed method, first, we make use of two known evaluation metrics: (a) \textit{play performance} \cite{Liu2018TowardID} of the online deep Q-network, and (b) \textit{fidelity} \cite{10.1145/3527448} of the mimic learner. Play performance measures how well the deep Q-network performs with the mimic learner estimating its temporal difference targets, while fidelity measures how well the mimic learner matches the predictions of the online deep Q-network. 

As far as play performance is concerned, we aim at minimizing the number of \textit{hotspots}, the \textit{average delay per flight} and the number of \textit{delayed flights}. As for fidelity, we use two metric scores: (a) the \textit{mean absolute error (MAE)} and (b) the \textit{accuracy} score. Given a minibatch of states, we calculate the MAE of this minibatch for any action as the mean absolute difference between the Q-values estimated by  the mimic learner and the Q-values estimated by the deep Q-network for that action. More formally, for a minibatch of states $D_s$, the $\text{MAE}_i$ of action $a_i$ is denoted as:

\begin{equation}
MAE_i = \frac{1}{|D_s|} \sum_{s \in D_s} | Q(s,a_i; \phi_X) - Q(s,a_i; \theta) |
\end{equation} 

\noindent
It is worth noting that minimizing the MAE of the mimic learner is very important for training XDQN. Since deep Q-network updates its parameters $\theta$ by using the mimic model to provide the target Q-values, large MAEs can lead deep Q-network to overestimate bad states and understimate the good ones, and thus, find very diverging policies that completely fail to solve the task. 

To calculate the accuracy score, again given a minibatch of states, for each state we compare the action selected by the mimic model and the online Q-network. Accuracy measures the percentage of the predictions of the two estimators that agree with each other, considering that both models select the action with the highest estimated Q-value. 

Second, we design appropriate experiments and illustrate XDQN's \textit{local} and \textit{global} interpretability. We focus  on providing aggregated interpretations, focusing on the contribution of features to local decisions and to the overall policy: This, as suggested by ATM operators,  is beneficial towards understanding decisions, helping them to increase their confidence to the solutions proposed, and mastering the inherent complexity in such a multi-agent setting, as solutions may be due to complex phenomena that are hard to be traced \cite{Kravaris2022ExplainingDR}.
Specifically, in this work, local explainability measures state features' importance on a specific instance (i.e. a single state-action pair), demonstrating which features contribute to the selection of a particular action over the other available ones. Global explainability aggregates feature importance on particular action selections over many different instances and {aims to explain the overall policy} of mimic learner. Third, we demonstrate global explainability of the DRL model through the whole training process, addressing the question of how a DRL model learns to solve the target task. 

\subsection{Experimental Scenarios and Settings}
Experiments were conducted on three in total scenarios. Each of these scenarios corresponds to a date in 2019 with heavy traffic in the Spanish airspace. In particular, the date scenarios, on which we assess our models, are 20190705, 20190708 and 20190714. However, to bootstrap the training process we utilize a deep Q-network  pre-trained in various scenarios, also including 20190705 and 20190708. In the training process, the deep Q-network is further trained according to the method we propose. The experimental scenarios were selected based on the number of hotspots and the average delay generated in the ATM system within the duration of the day, which shows the difficulty of the scenario. We note that for each scenario we ran five separate experiments and average results. 

Table \ref{tab:table1} presents information on the three experimental scenarios. In particular, the flights column indicates the total number of flights (represented by agents) during the specific day. The initial hotspots column indicates the number of hotspots appearing in the initial state of the scenario. The flights in hotspots column indicates the number of flights in at least one of the initial hotspots. Note that all three scenarios display populations of agents of similar size, with 20190708 having the smaller population and the least initial hotspots.

\begin{table}[t]
\centering
 \begin{tabular}{c c c c} 
 \hline
 Scenario & Flights & Initial Hotspots & Flights in Hotspots \\ [0.5ex] 
 \hline
 20190705 & 6676 & \textbf{100} & \textbf{2074} \\ 
 20190708 & 6581 & 79 & 1567 \\
 20190714 & \textbf{6773} & 92 & 2004 \\
 \hline
 \end{tabular}
 \caption{The three experimental Air Traffic Management (ATM) scenarios} 
 \label{tab:table1}
\end{table}

\subsection{Implementation Details}


In our implementation setting we utilize a deep multilayer perceptron as the Q-network. 
In particular, we use an $\epsilon$-greedy policy, which at the start of exploration has $\epsilon$ equal to 0.9 decaying by 0.01 every 15 episodes until reaching the minimum of 0.04. The total number of episodes are set to 1600 and the update target frequency is set to 9 episodes. In the exploitation mode, we set $\epsilon$ equal to 0.04. We set the maximum depth of the Gradient Boosting Regressor equal to 45 and the number of minimum samples for a split equal to 20. We also use the mean squared error as the splitting criterion. To train a single decision tree for all different actions, we create a non binary splitting rule of the root based on the action size of the task, so that the state-action pairs sharing the same action match the same subtree of the splitting root. Empirically, we set the memory capacity of the experience replay for the mimic learner, i.e. the hyperparameter $K$, equal to the $1/20$ of the product of three other hyperparameters, namely the total number of timesteps per episode (set to 1440), the update target frequency (set to 9) and the number of agents (set to 7000). Thus, $K$ is set to 4536000 steps.

\subsection{Evaluation of play performance}

\begin{table*}[h!]
\setlength\tabcolsep{30pt} 
\begin{tabular*}{\textwidth}{cccc}
\toprule 
Action (Delay Option) & \multicolumn{3}{c}{XDQN mimic models}\\
\cmidrule(lr){2-4} 
& X0705 & X0708 & X0714 \\
\midrule 
0    & 0.279 & 0.237 & 0.291 \\ 
1    & 1.766 & 1.971 & 1.942   \\
2    & 0.910 & 0.928 & 1.002  \\
3    & 0.575 & 0.661 & 0.640  \\
4    & 0.639 & 0.748 & 0.725  \\
5    & 1.893 & 2.096 & 2.121  \\
6    & 1.590 & 1.766 & 1.715  \\
7    & 1.610 & 1.816 & 1.733  \\
8    & 0.449 & 0.514 & 0.497  \\
9    & 0.740 & 0.849 & 0.823 \\
10   & 1.292 & 1.525 & 1.461  \\
\bottomrule 
\end{tabular*}
\caption{Evaluation of the average Mean Absolute Errors (MAE) of the trained mimic models over all mimic updates}\label{tab:table4}
\end{table*}
\normalsize


\begin{figure}[t]
\includegraphics[scale=.38]{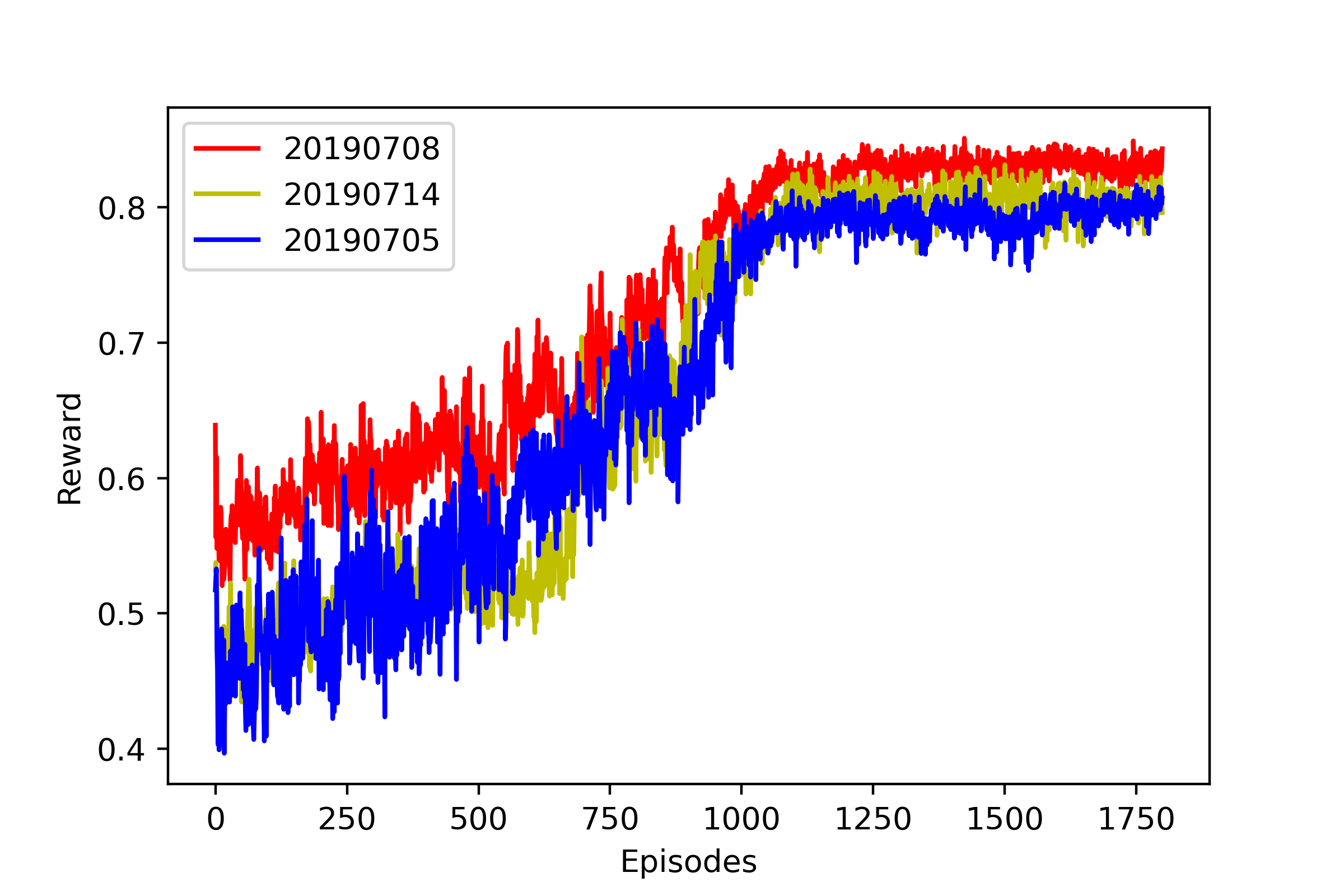}
\caption{Episodic reward in the three evaluated ATM scenarios}
\label{rewards.png}
\end{figure}

Table \ref{tab:table3} demonstrates the performance of DQN and XDQN on the three experimental scenarios. The final hotspots column indicates the number of unresolved hotspots in the final state: It must be noted that these hotspots may have emerged due to delays assigned to flights and may be different than the hotspots at the beginning of each scenario.  The average delay per flight column shows the total minutes of delay imposed, divided by the number of flights in the specific scenario. The delayed flights column indicates the number of flights affected by more than four minutes of delay, as it is done by operators.

We observe that XDQN performs similar to DQN in all three evaluated metric scores. In particular, DQN slightly outperforms XDQN in terms of the final hotspots and average delay in all three scenarios. Nonetheless, XDQN achieves to decrease the number of the delayed flights in one scenario, while it demonstrates competitive performance on the others. Figure \ref{rewards.png} shows the episodic reward of XDQN over time: XDQN manages to reach convergent behavior in all scenarios by retaining high episodic rewards.

\subsection{Evaluation of fidelity}

As discussed in Subsection 4.2, for the fidelity evaluation, we measure the mean absolute error (MAE) and the accuracy score. Given the DCB experimental scenarios, we train three different mimic models; namely X0705, X0708 and X0714.  Table \ref{tab:table4} reports the average MAE for each decided action over all mimic learning updates. We observe that all errors range in relatively small quantities, given that in testing, the absolute Q-values hovered around 200. As we highlighted above, this is very important for stabilizing the training process of XDQN, since we need very accurate mimic Q-value predictions, ideally equal to the ones generated by the deep Q-network. 

To further assess the fidelity of XDQN mimic learner, Table \ref{tab:table5} illustrates the average accuracy scores over all mimic learning updates. Since a Gradient Boosting Regressor mimic learner is a boosting algorithm, it produces sequential decision trees that can successfully seperate the state space and approximate well the predictions of the deep Q-network function. We observe that the mimic learner and the deep Q-network agree with each other to a very good extent; namely from approximately 81\% to 91\%. Therefore, we expect the mimic learner to be able to accumulate the knowledge from the deep Q-network with high fidelity.

\begin{table}[t]
\centering
 \begin{tabular}{c c } 
 \hline
 Scenario & Accuracy (\%) \\ [0.5ex] 
 \hline
 20190705 & 88.45  \\ 
 20190708 & 81.89 \\
 20190714 & 90.88 \\
 \hline
 \end{tabular}
 \caption{The accuracy scores of mimic models} 
 \label{tab:table5}
\end{table}

\subsection{Local and Global Explainability}

\begin{table*}[t]
\setlength\tabcolsep{20pt} 
\begin{tabular*}{\textwidth}{ccc}
Feature Index & Feature Meaning & ACD  \\
\midrule 
0    & Delay the corresponding flight has accumulated up to this point & Positive  \\ 
1    & Total number of hotspots the corresponding flight participates in
 & Positive  \\
3    & The sector in which the second hotspot the corresponding flight participates occurs & Positive  \\
63   & The minute of day the flight takes off given the delay (CTOT) & Negative \\
64   & The minutes the flight remains in the first sector it crosses & Negative  \\
68   & The minutes the flight remains in the fifth sector it crosses  & Negative  \\
\bottomrule 
\end{tabular*}
\caption{Demonstration of the most significant state features in terms of average contribution difference (ACD) in selecting the no-delay action versus a delay action. A positive ACD means that the corresponding state feature on average contributes more to the selection of the no-delay action ``0". On the contrary, a negative ACD means that the corresponding state feature on average  contributes  to the selection of a delay action ``1 - 10".}\label{tab:table6}
\end{table*}

In the DCB setting, it is important for the operator to understand how the system reaches decisions on regulations (i.e. assignment of delays to flights): This, as already pointed out, should be done at a level of abstraction that would allow them to increase their confidence to the solutions proposed, mastering the inherent complexity of the setting. Therefore, we are mainly interested in receiving explanations about which state features contribute to the selection of delay actions over the no-delay action (i.e. action equal to 0).  

\begin{figure}[t]
\includegraphics[scale=.32]{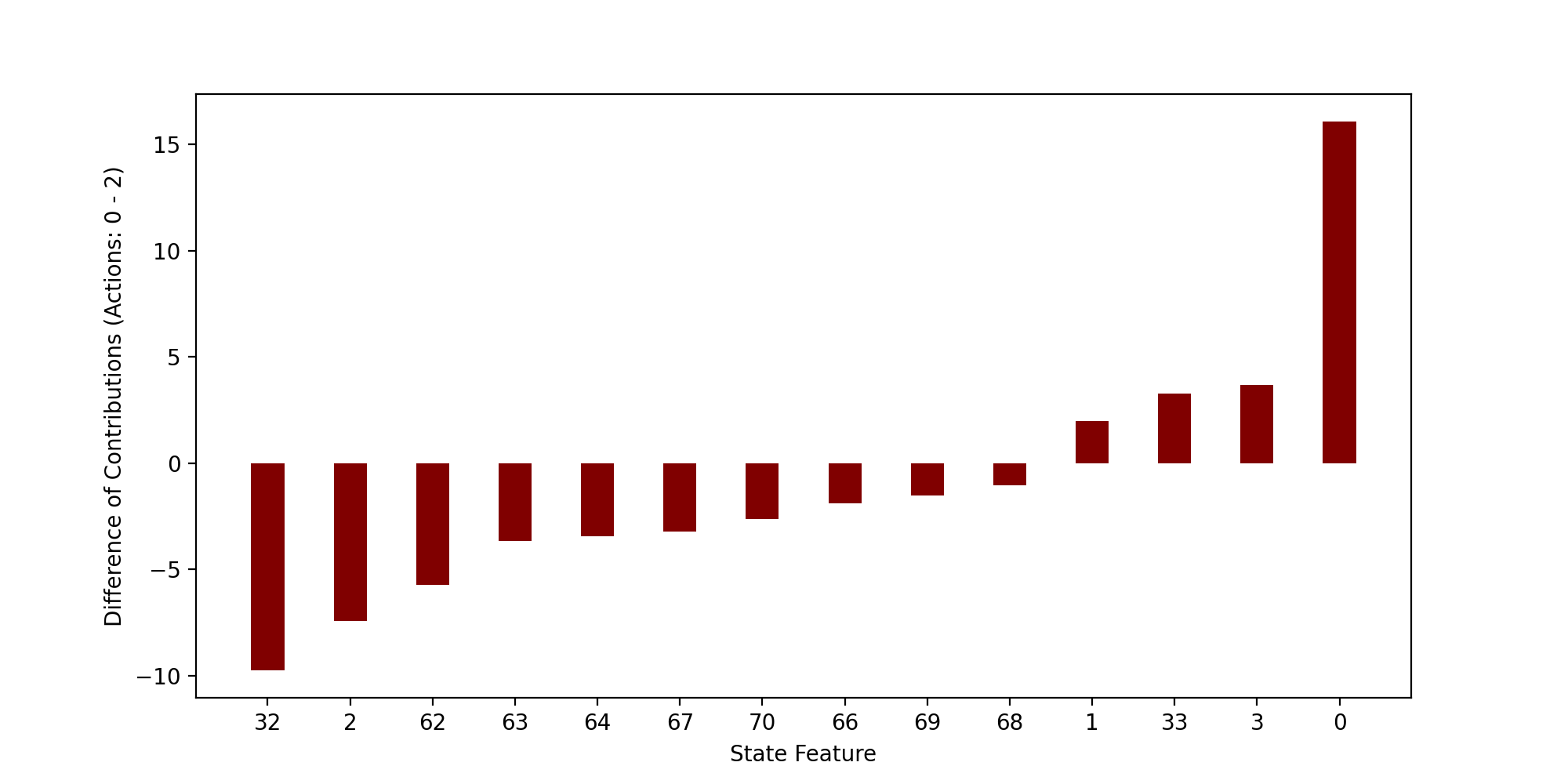}
\caption{Illustration of significant differences of feature contributions to Q-value in selecting action "0" and action "2" in a single state, in which action "2" was selected. Positive differences mean that the respective state features have a greater contribution to Q-value when action "0" is selected, rather than when action "2" is selected. Negative differences have the opposite meaning.}
\label{local_2.png}
\end{figure}

First, we demonstrate the ability of the mimic learner to provide local explainability. As already said, local explainability involves showing which state features contribute to the selection of a particular action over the other available ones in a specific state. To this aim, we work on pairs of actions - let $a_1$ and $a_2$ - and calculate the differences of feature contributions in selecting $a_1$ and $a_2$ in a single state. To highlight only the most significant differences, we focus only on those features whose differences are above a threshold. Empirically, we set this threshold equal to 0.5. Figure \ref{local_2.png} illustrates local explainability on a given state in which action "2" was selected.  Figure \ref{local_2.png} provides the differences of feature contributions to the estimation of Q-values when selecting action "0" against selecting action "2"  (denoted by ``0-2"). We observe that the features that contributed more to the selection of the delay action "2" were those with index 32 (i.e. The sector in which the last hotspot occurs), 2 (i.e. the sector in which the first hotpot occurs) and 62 (i.e. the minutes that the flight spends crossing the last sector).

Finally, we demonstrate XDQN global explainability by aggregating the importance of features on particular action selections over many different state-action instances. In particular, we are interested in measuring the state feature contributions to the selection of delay actions (i.e. actions in the range $[1, 10]$) over the no-delay action (i.e. action "0") in the overall policy. To this aim, we work on all possible pairs of actions, with one action always being the no-delay action and the other one being a delay action, and average the differences of feature contributions to estimating the Q-value in selecting those actions over many different state-action instances with the same selected delay action. Table \ref{tab:table6} shows the most significant state features in terms of average contribution difference (ACD) in selecting the no-delay action versus a delay action. To select those features, we initially filter the most significant ones, namely the features whose absolute ACD is greater than a threshold, for each action in the range $[1, 10]$) over the no-delay action (i.e. action "0"), and present the three most common features with positive and negative ACD. We observe that features with index 0, 1 and 3 contribute more to the selection of the no-delay action. On the contrary, features with indexes 64, 63 and 68 contribute more to the selection of a no-delay action.

\begin{figure}[t]
        \centering
        \begin{subfigure}{\linewidth}
            \centering
            \includegraphics[scale=.45]{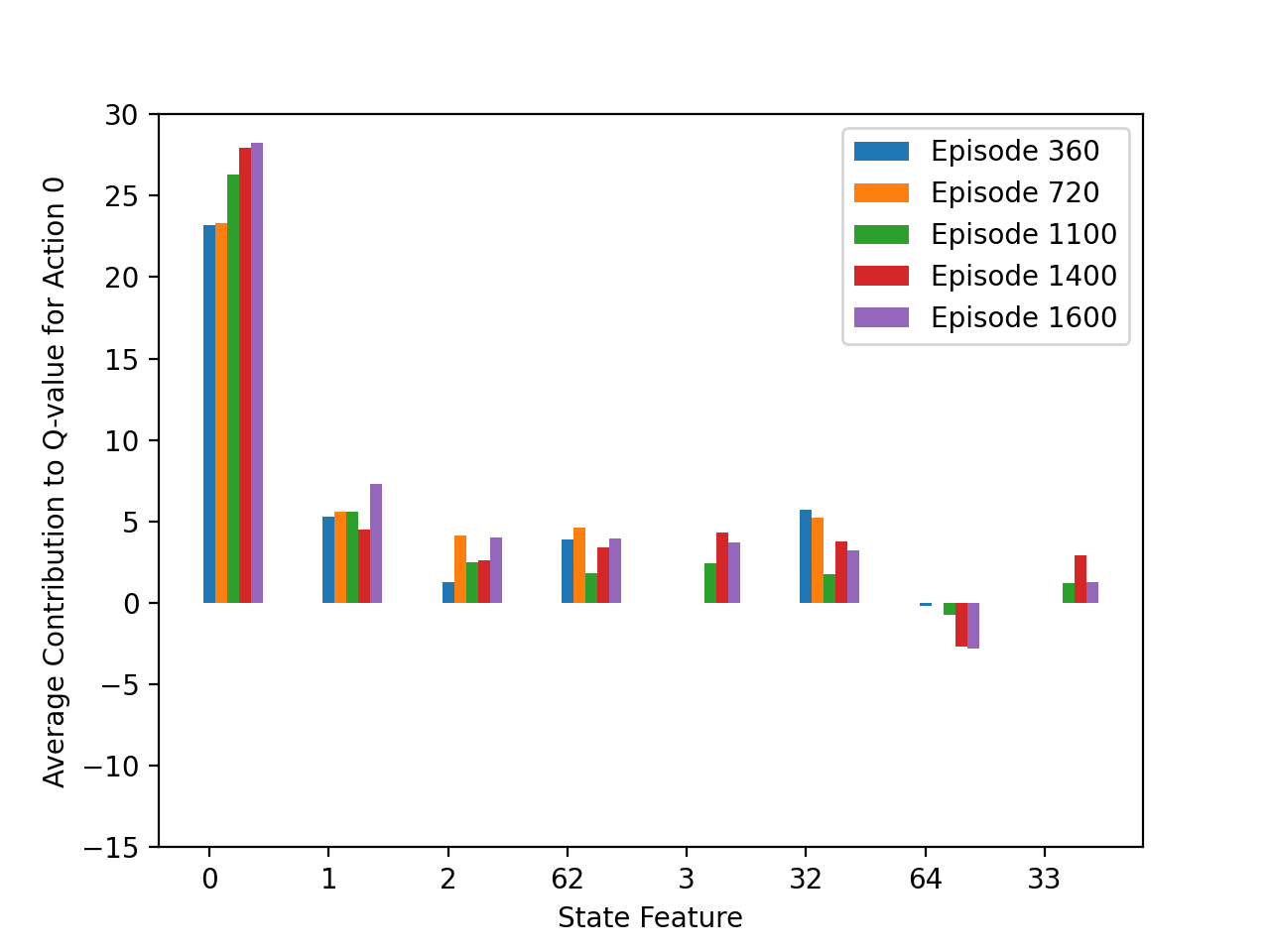}
        \end{subfigure}
        \centering
        \begin{subfigure}{\linewidth}
            \centering
            \includegraphics[scale=.45]{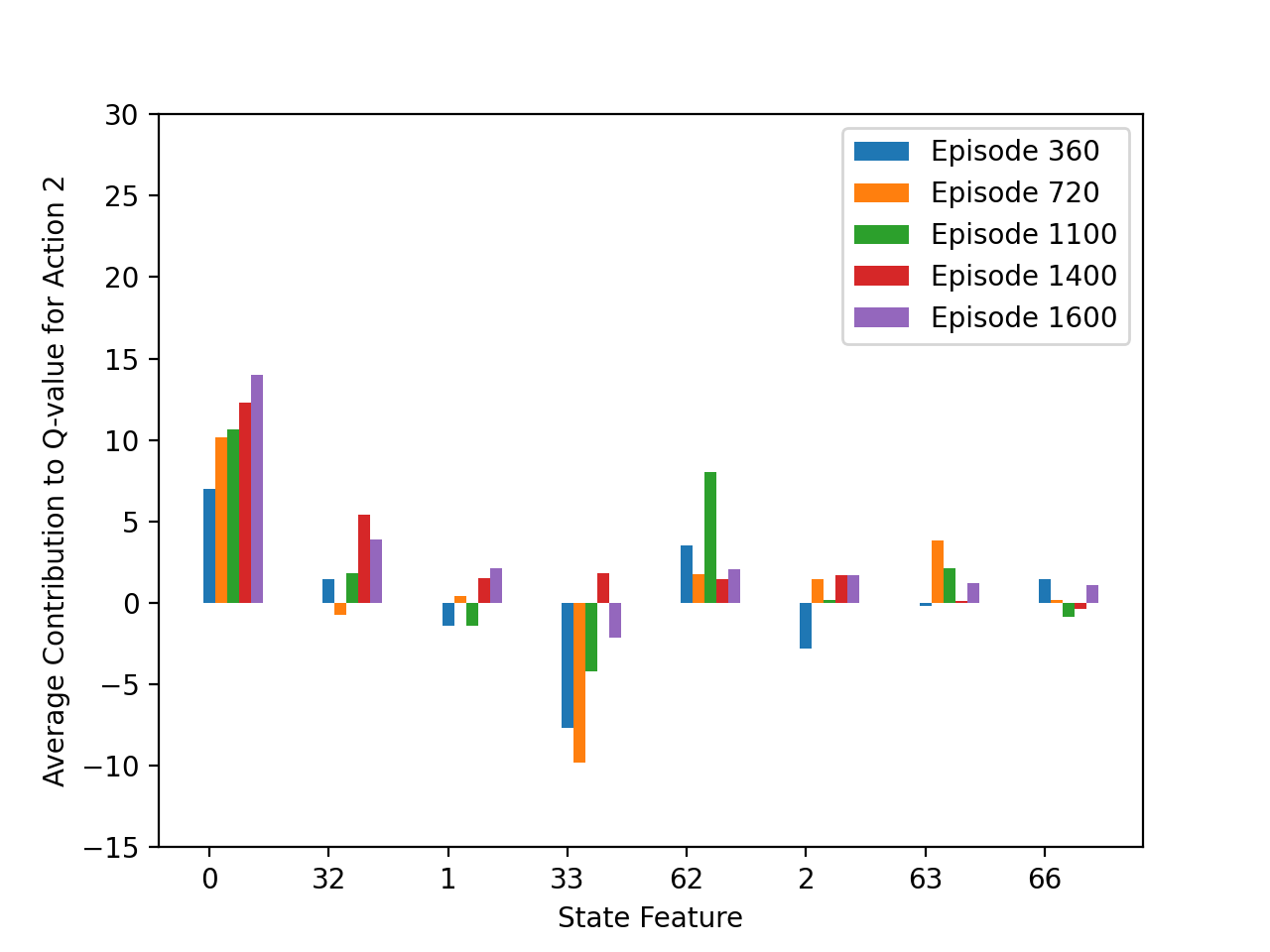}
        \end{subfigure} 
        \caption{Illustration of the evolution of features' contributions for selecting the no-delay action ("0") and a delay one ("2") through 5 representative training episodes (360th, 720th, 1100th, 1400th and 1600th) in terms of absolute average feature contribution (AAFC) to Q-value for the eight features with highest AAFC values in the final model (episode 1600) in the selection of the aforementioned actions.}
        \label{fig:old_instances}
\end{figure}

Last but not least, we demonstrate how global explainability evolves through the training process, addressing the question of how a DRL model learns to solve the target task. To this aim, we measure the absolute average feature contribution (AAFC) to Q-value at different training episodes for the selection of each action. Figure \ref{fig:old_instances} illustrates the evolution of global explainability for selecting the no-delay action and a delay action through 5 representative training episodes (360th, 720th, 1100th, 1400th and 1600th), in terms of AAFC to Q-value for the eight features with highest AAFC values in the final model (episode 1600) in the selection of the aforementioned actions. We observe that for both evaluated actions most of the features show an increasing/decreasing trend in their average contribution to Q-value over time, such as those with indices 0, 1 and 63. It is worth noting that although the features with indexes 0 and 1 have been highlighted as the most significant for the selection of the no-delay action, they have also significant but less contribution to a delay action as well. 

\section{Related Work}
Explainability in Deep Reinforcement Learning (DRL) is an emergent area whose necessity is related to the fact that DRL agents solve sequential tasks, acting in the real-world, in operational settings where safety, criticality of decisions and the necessity for transparency (i.e. explainability with respect to real-world pragmatic constraints \cite{10.1145/3527448}) is the norm. However, DRL methods use closed-boxes whose functionality is intertwined and are not interpretable: This may hinder DRL methods explainability.  In this paper we address this problem by proposing an interpretable DQN method comprising two models which are trained jointly: An interpretable mimicking model and a deep policy model. The later offers training samples to the mimicking one and the former interpretable model offers target action values for the other to improve its predictions. At the end of the training process, the mimicking model has the capacity to provide  high-fidelity interpretations to the decisions of the deep policy model.
This is a specific example for interpreting DRL methods, according to the interpretable box design paradigm: This paradigm follows the conjecture (stated for instance in \cite{InterpretableML})  that there is high probability that the accuracy of closed boxes can be approximated by well designed interpretable models. In this work, following this paradigm, we train an interpretable  model via mimicking, in parallel to the online Q network. Distillation could be another option \cite{PolicyDistillation}, but in this work we explore mimicking as a process to train inherently interpretable models, such as decision trees. 

There are many  proposals for interpreting deep NNs models, through  distillation and mimicking approaches. These approaches differ in several dimensions: (a) the targeted representation (e.g., decision trees in DecText \cite{ExtractingDTfromTrainedNNs}, logistic model trees (LMTs) in reference \cite{LMTExtractionfromANNs}, or Gradient Boosting Trees in reference \cite{Che2016InterpretableDM}), (b) to the different splitting rules used towards learning a comprehensive representation, (c) to the actual method used for building the interpretable model (e.g., \cite{LMTExtractionfromANNs} uses the LogiBoost method, reference \cite{ExtractingDTfromTrainedNNs} proposes the DecText method, while the approach proposed in reference \cite{Che2016InterpretableDM} proposes a pipeline with an external classifier, (d) on the way of generating samples to expand the training dataset. 
These methods  can be used towards interpreting constituent individual DRL models employing (deep) NNs. The interested reader is encouraged to read a thorough review on these methods provided in \cite{Survey, Murdoch22071, PrinciplesPracticeofXML, InterpretableML}.

For DRL, authors in \cite{Liu2018TowardID} introduce Linear Model U-trees (LMUTs) to approximate predictions for DRL agents. 
An LMUT is learned by an on-line
algorithm that is well-suited for an active play setting.  The use of LMUTs is compared against using  CART, M5 with regression tree, Fast Incremental Model Tree (FIMT) and with Adaptive Filters (FIMT-AF).
The use of decision trees as interpretable policy models trained through mimicking has been also investigated in \cite{DistalExplanations}, in conjunction to using a causal model representing agent's objectives and opportunity chains. However, the decision tree in this work is used to infer the effects of actions approximating the causal model of the environment. Similarly to what we do here, the decision tree policy model is trained concurrently with the RL policy model, assuming a model-free RL algorithm and exploiting state-action samples using an experience replay buffer. 
In \cite{DistillingDRLInSDTs} authors illustrate
how Soft Decision Trees (SDT)  \cite{DistillingNNsInSDTs} can
be used in spatial settings as interpretable policy models. SDT are hybrid classification models of 
binary trees of predetermined depth, and  neural networks.  However their inherent interpretability is questioned given their structure.
Other approaches train interpretable models other than trees, such as the Abstracted Policy Graphs (APGs) proposed in \cite{PolicyLevelExplanations}, assuming a well-trained policy model. APGs  can offer interpretable representations of policies, concisely summarizing them, so that individual decisions can be explained
in the context of expected future transitions. 

Approaches following the interpretable box design paradigm also use use attention models for visual agents \cite{betterInterpretability, AttAugmentedAgents}, and interpretable policy models in a rather direct way \cite{Pyeatt, ConservativeQImprovement}.

In contrast to the above mentioned approaches, XDQN  can be applied to any setting with arbitrary state features, where the interpretable model formed using Gradient Boosting Regressors is trained jointly to a deep one through mimicking in an active play setting, following the DQN algorithm. 
It is worth noting that experimentally, instead of Gradient Boosting Regressors, we also tested naturally interpretable Linear Trees (such as LMUTs \cite{Liu2018TowardID}); i.e. decision trees with linear models in their leaves). However, such approaches completely failed to solve the task, demonstrating quite low play performance with very large mean absolute errors.

As far as explanations are concerned, we opted for features' contributions to the Q-values, in a rather aggregated way, using the residue of each Gradient Boosting Regressor node, as done in \cite{DELGADOPANADERO2022199}. This approach, as shown in \cite{DELGADOPANADERO2022199}, reports advantages over using well known feature importance calculation methods, avoiding linearity assumptions made by LIME \cite{10.1145/2939672.2939778} and bias in areas where features have high variance, and also avoiding taking all tree paths into account in case of outliers, as done by SHAP \cite{lundberg2017unified}.

\section{Conclusion and Future Work}

In this work, we address the challenging issue of training interpretable policy models for solving real-world problems, such as the multi-agent demand-capacity balancing problem pertaining to air traffic management. To this aim, we have trained interpretable deep Q-learning models through mimic learning without requiring the existence of already well-trained deep Q-networks. Experimentally, we have shown that the proposed interpretable  XDQN method, utilizing a Gradient Boosting Regressor as the mimic learner, performs on a par with DQN in terms of play performance whereas demonstrating high fidelity.

Further work on XDQN is to design, evaluate and compare various explainable mimic models that can effectively substitute the target Q-Network. Moreover, the proposed mimicking paradigm is generic, and can be naturally extended to many well-known DRL algorithms. Thus, future steps should also aim at benchmarking our methodology utilizing state-of-the-art DRL in various experimental settings.

\begin{acks}
Acknowledgements will appear in the final version of this manuscript.
\end{acks}
 
\bibliographystyle{ACM-Reference-Format} 
\bibliography{references.bib}

\end{document}